\pdfoutput=1

\documentclass[11pt]{article}

\usepackage{EMNLP2023}

\usepackage{times}
\usepackage{latexsym}

\usepackage[acronym, hyperfirst=false]{glossaries}
\usepackage{listings}
\usepackage{xcolor} 

\usepackage[T1]{fontenc}

\usepackage[utf8]{inputenc}

\usepackage{microtype}

\usepackage{inconsolata}

\usepackage{array,multirow}
\usepackage{enumitem}
\usepackage{booktabs}
\usepackage{graphicx}
\usepackage{makecell}
\usepackage{spverbatim}
\usepackage{textcomp}  
\usepackage{xspace}

\newcolumntype{H}{>{\setbox0=\hbox\bgroup}c<{\egroup}@{}}

\usepackage[loadshadowlibrary,disable]{todonotes}
\setuptodonotes{color=blue!10, linecolor=black, shadow}

\title{MetricX-25 and GemSpanEval: Google Translate\\Submissions to the WMT25 Evaluation Shared Task}

\author{
  Juraj Juraska, Tobias Domhan, Mara Finkelstein, Tetsuji Nakagawa, \\
  \textbf{Geza Kovacs, Daniel Deutsch, Pidong Wang, \and Markus Freitag} \\
  Google Translate \\
  \texttt{\{jjuraska,domhant,freitag\}@google.com}
}

\defcitealias{gemma3technicalreport}{Gemma Team, 2025}

\defcitealias{gemini25technicalreport}{Gemini Team, 2025}

\newacronym{llm}{LLM}{Large Language Model}
\newacronym{mqm}{MQM}{Multidimensional Quality Metrics}
\newacronym{qe}{QE}{quality estimation}
\newacronym{rl}{RL}{reinforcement learning}

\begin{document}

\maketitle

%
%

\begin{abstract}
In this paper, we present our submissions to the unified WMT25 Translation Evaluation Shared Task.
For the Quality Score Prediction subtask, we create a new generation of MetricX with improvements in the input format and the training protocol, while for the Error Span Detection subtask we develop a new model, GemSpanEval, trained to predict error spans along with their severities and categories.
Both systems are based on the state-of-the-art multilingual open-weights model Gemma~3, fine-tuned on publicly available WMT data.
We demonstrate that MetricX-25, adapting Gemma~3 to an encoder-only architecture with a regression head on top, can be trained to effectively predict both MQM and ESA quality scores, and significantly outperforms its predecessor.
Our decoder-only GemSpanEval model, on the other hand, we show to be competitive in error span detection with \textsc{xComet}, a strong encoder-only sequence-tagging baseline.
With error span detection formulated as a generative task, we instruct the model to also output the context for each predicted error span, thus ensuring that error spans are identified unambiguously.
\end{abstract}

\section{Introduction}
\label{sec:introduction}

Large language models~(LLMs) have been evolving at a breakneck speed over the past couple of years, achieving performance in machine translation~(MT) that matches or even exceeds that of humans for certain languages and domains~\citep{kocmi-etal-2024-findings}.
While human evaluation is still the most reliable way to assess the quality of MT models, in this fast-paced environment of state-of-the-art models being released on a monthly basis, the cost and duration make it infeasible to run human evaluation studies to benchmark models regularly.
Improving automatic metrics is therefore instrumental to making further progress in MT, especially as the field expands to low-resource languages and more difficult domains.


Currently, the most successful automatic MT evaluation metrics are trained neural models themselves, following one of two main paradigms: (1)~regression models predicting a scalar quality score~\citep{rei-etal-2022-comet,juraska-etal-2024-metricx}, and (2)~sequence-tagging or generative models, providing fine-grained quality feedback, including error spans, severities and categories~\citep{fernandes-etal-2023-devil,guerreiro-etal-2024-xcomet}.
In this work, we push the performance of automatic metrics in both categories higher by leveraging a state-of-the-art multilingual open-weights LLM, Gemma~3~\citep{gemma3technicalreport}, resulting in two separate submissions to the WMT25 Evaluation Shared Task.


For the Quality Score Prediction subtask, we develop \emph{MetricX-25}, a successor to MetricX-24~\citep{juraska-etal-2024-metricx}, updated to use an encoder-only architecture and trained on a combination of publicly available direct assessment (DA) and \gls*{mqm} scores from WMT shared tasks between 2015 and 2023.
For the Error Span Detection subtask, we introduce \emph{GemSpanEval}, a generative model that identifies and categorizes error spans in JSON format, trained exclusively on \gls*{mqm} error span annotations from WMT20--24.
We enable GemSpanEval to uniquely identify short, non-unique error spans by training the model to also indicate the error span context where necessary.

The key takeaways from our experiments, detailed in this report, include:
\begin{enumerate}[itemsep=-0.25em]
    \item Gemma 3 offers a strong multilingual foundation for an automatic MT evaluation metric, and adapting it to an encoder-only architecture proves highly effective for score prediction.
    \item It is possible to train an automatic metric to effectively predict different types of score using a single regression head by simply mixing training examples of different scores, with a score type indication included in the input.
    \item Fine-tuning a strong multilingual decoder-only model can be competitive with encoder-only error span detection models.
    \item Predicting error span context can be used to uniquely identify error spans when formulating span detection as a generative task.
\end{enumerate}
 
\section{Data}
\label{sec:data}

The systems we developed for both the quality score prediction and the error span detection are trained solely on publicly available data from the WMT Metrics shared tasks between 2015 and 2024.
Quality ratings for system translations across those 10 years were collected using 3 different methods:
\begin{enumerate}
    \item Direct assessment (DA) scores, provided mostly by non-expert raters, on a scale from 0 to 100. WMT data with DA scores is available for years between 2015 and 2023, and covers nearly 50 language pairs.
    \item MQM scores~\citep{lommel2014multidimensional,freitag-etal-2021-experts} on a scale from 0 to 25 (or uncapped in the most recent years), where lower is better. The scores are derived from professional translator annotations of error spans, including error severities and categories, where, generally, each minor error and each major error contribute 1 and 5 to the score, respectively. MQM annotations are only available for a limited number of language pairs (en-de, en-es, en-ru, en-zh, he-en, zh-en, ja-zh) for WMT data starting from 2020.
    \item ESA scores~\citep{kocmi-etal-2024-error}, which combine the first two approaches in that the expert raters annotate error spans and severities, yet they provide an overall quality score on a scale from 0 to 100 as well. These were only introduced in WMT24, so there is only one year worth of ESA data.
\end{enumerate}
We use both DA and MQM scores for training MetricX, our score prediction system, and MQM error span annotations for training GemSpanEval.
In general, we reserved data from WMT24 -- both MQM~\citep{freitag-etal-2024-llms} and ESA~\citep{kocmi-etal-2024-findings} -- for validation of our models, with the exception of one submission to the error span detection task.
We provide more details on the training and validation sets in~\S\ref{sec:metricx} for MetricX and in~\S\ref{sec:gemspaneval} for GemSpanEval.

\section{Quality Score Prediction: \textit{MetricX-25}}
\label{sec:metricx}

Our MetricX-25 submissions to the Quality Score Prediction subtask are based on the successful MetricX-24~\citep{juraska-etal-2024-metricx, freitag-etal-2024-llms}, with several modifications and improvements.
The biggest one among them is the switch from mT5~\citep{xue-etal-2021-mt5} to Gemma~3 as the backbone model.
We start this section by providing an overview of the similarities and differences between MetricX-24 and MetricX-25, then give more details on the training and evaluation data, and finally describe our experiments and the MetricX-25 systems we submitted to the shared task.

\subsection{MetricX Overview}

MetricX is a regression model trained to predict a quality score for a machine translation given the source segment and/or a reference translation.
As of the `24 version, there are no separate models for reference-based and reference-free prediction; instead, a single model is trained on a mixture of examples: (1) with both the source and a reference, (2) with the reference omitted, and (3) with the source omitted.

The model is trained on translation evaluation data in two stages.
It is first fine-tuned on $z$-normalized DA scores, then further fine-tuned on a mixture of MQM scores and raw DA scores.
In both stages, a small proportion of synthetic training data, generated from WMT data, is included to help MetricX models recognize certain types of bad translations that are insufficiently represented in the standard WMT data.
These synthetic examples cover cases such as over- and undertranslation, fluent but unrelated translation, and missing punctuation.
We refer the reader to~\citet{juraska-etal-2024-metricx} for the full list of synthetic example categories and details on how the data was constructed.

\subsection{What Is New in MetricX-25?}

\paragraph{Initialization model.} The model we initialize MetricX-25 from is Gemma~3 12B, a state-of-the-art multilingual open-weights model similar in size to the 13B-parameter mT5-XXL used in all previous versions of MetricX.
In contrast to mT5-based MetricX models, MetricX-25 uses an encoder-only architecture with a regression head on top.
Specifically, MetricX-25 is a fine-tuned Gemma Encoder~\citep{suganthan2025adapting} with mean pooling and no uptraining, and with the encoder's weights initialized from the corresponding decoder weights of Gemma~3.
Besides the major differences in architectures and pretraining corpora and strategies, Gemma~3 also has the advantage of supporting significantly longer context windows (up to 128K tokens) than mT5 and has an even wider language support (over 140 languages, compared to mT5's 101).

\begin{figure}
    \small
    \centering
    \begin{tabular}{m{0.94\linewidth}}
        Czech source: \\
        \textasciigrave\textasciigrave\textasciigrave Připadalo mi, že na mě dýchnul závan z hrobu.\textasciigrave\textasciigrave\textasciigrave \\
        \\
        English (United Kingdom) reference: \\
        \textasciigrave\textasciigrave\textasciigrave It was like having felt a draught from a grave.\textasciigrave\textasciigrave\textasciigrave \\
        \\
        English (United Kingdom) translation: \\
        \textasciigrave\textasciigrave\textasciigrave It was like having felt a draft from a grave.\textasciigrave\textasciigrave\textasciigrave \\
        \\
        Score type: MQM \\
    \end{tabular}
    \caption{Example MetricX-25 model input.}
    \label{fig:metricx_example_input}
    \vspace{-0.1in}
\end{figure}

\paragraph{Language indication.} For MetricX-25, we augment the model input with source and target language information.
The motivation behind this is to help MetricX recognize when an untranslated source (or portion of it) is appropriate, as well as help it handle quality assessment of translations from one language dialect to another without incorrectly assuming the translation is largely untranslated.
Thus, we include the country information too if the locale is indicated in the data (e.g., ar\_EG) and the language has multiple major dialects, such as Arabic (Egyptian, Modern Standard Arabic, etc.) or Portuguese (Brazilian, European, etc.).
Figure~\ref{fig:metricx_example_input} shows a full example of a model input.

\paragraph{Input format.} Given the dual nature of human evaluation in the WMT25 shared task -- MQM for some language pairs and ESA for others -- we also add a score type indication to the model input, so that it learns to predict both types of quality score.
We use the ``MQM'' score type for the MQM training and evaluation data, and ``ESA'' for the DA training and ESA evaluation data.
Considering the possibly multi-paragraph segments in the official test sets, we also enclose each segment between triple backticks and separate input segments with double newline characters (see Figure~\ref{fig:metricx_example_input}).\footnote{We experimented with including a preamble with instructions on the quality assessment task in the input too, but it did not improve the performance, perhaps except for the first few steps of fine-tuning.}

\paragraph{2-way hybrid input mode.} We changed the training recipe for MetricX-25 by only including reference-only examples in the first stage of fine-tuning.
In the second stage, we fine-tune the model on two types of examples only: (1) source-only, and (2) with source and reference both.
The reason is twofold.
First, MQM scores were produced in a source-based fashion, without a reference being available at all, so training examples with an MQM score but no source segment might not provide an accurate signal to the model.
We verified experimentally that omitting these examples in the second stage does not have a significant negative impact on reference-only prediction performance.
Second, in all the evaluation scenarios of the quality score prediction task, source segments are available, so there is little reason to distract the model with source-free training examples in the second stage of fine-tuning.

\paragraph{Score clipping.} Earlier versions of MetricX had MQM scores in the training data, as well as the output scores, clipped to the [0, 25] range.
However, with the switch to document-level segments over the past two years of the shared task, and the fact that MQM scores in human evaluation are therefore no longer capped at 25, we also drop the score clipping in MetricX-25, allowing for output scores greater than 25, which we expect to improve the correlation with MQM scores for long segments.\footnote{This causes a small discrepancy with the synthetic training data, which uses a fixed range of [0, 25], with many of the examples having a score of 25 assigned to them. Nevertheless, most of the synthetic examples are sentence-level, so an MQM score of 25 is reasonable for very bad translations.}

\subsection{Experimental Setup}

\paragraph{Training data.} In the first stage of fine-tuning MetricX-25, we use $z$-normalized DA scores from WMT15--23, with the into-English subset of WMT21 omitted due to its low quality~\citep{juraska-etal-2023-metricx}.
Furthermore, the DA $z$-scores are aggregated per segment, negated, and finally clipped to the $[-1.0, 1.0]$ range, as shown by~\citet{juraska-etal-2023-metricx} to yield the best performance.
We also incorporate a small proportion of the synthetic training data introduced in~\citet{juraska-etal-2024-metricx}.
In this stage, we do not include any score type indication in the input.
In the second fine-tuning stage, we mix an equal proportion of the same DA data as above (with ``ESA'' indicated as the score type) and MQM data from WMT20--23 (with ``MQM'' score type), along with synthetic data included in each group of examples.
In contrast to the first stage, however, we use raw DA scores rescaled to the MQM scale here, so the model does not have to learn two different scales, only distributions.
We then rescale the output scores to their respective ESA and negative MQM scales, as expected for the evaluation on the official test set, in postprocessing.

\paragraph{Meta-evaluation.} Our validation set, which we use to pick the best model checkpoints, consists of both the MQM and ESA data from WMT24.
To evaluate our models' performance, we calculate to what degree their predicted scores agree with the human judgments of translation quality.
For segment-level correlation we use the tie-calibrated pairwise accuracy introduced by~\citet{deutsch-etal-2023-ties}, while at the system level we calculate soft pairwise accuracy~(SPA; \citealt{thompson-etal-2024-improving}).
These were the two primary meta-evaluation metrics used in the WMT24 Metrics Shared Task~\citep{freitag-etal-2024-llms}.
We use the same checkpoint selection strategy as for MetricX-24, averaging over all three MQM language pairs of WMT24, and downweighting the system-level component due to its larger variance.

\paragraph{Implementation details.} MetricX-25 is implemented in TensorFlow~\citep{tensorflow2015-whitepaper}, and all of our submitted MetricX-25 systems are based on the Gemma 3 variant with 12B parameters.
We defer further implementation details to Appendix~\ref{app_sec:metricx_implementation_details}.

\subsection{Results and Submission Details}

Here we present the results of our experiments, focusing on assessing the impact of the combined MQM/ESA score prediction and comparing the performance of the Gemma-based MetricX-25 with that of the similarly-sized mT5-based MetricX-24. Due to limited resource availability, we were only able to run each experiment with one random seed.

\subsubsection{Combining MQM/ESA Score Prediction}

\begin{table*}
    \small
    \centering
    \begin{tabular}{>{\centering\arraybackslash} m{0.18\linewidth}  >{\centering\arraybackslash} m{0.07\linewidth} >{\centering\arraybackslash} m{0.07\linewidth} >{\centering\arraybackslash} m{0.07\linewidth} >{\centering\arraybackslash} m{0.08\linewidth} >{\centering\arraybackslash} m{0.07\linewidth} >{\centering\arraybackslash} m{0.07\linewidth} >{\centering\arraybackslash} m{0.07\linewidth} >{\centering\arraybackslash} m{0.08\linewidth}}
        \toprule
        \multirow{2}{*}{\textbf{\makecell{Training protocol}}} & \multicolumn{4}{c}{\textbf{Segment-level pairwise accuracy}} & \multicolumn{4}{c}{\textbf{System-level soft pairwise accuracy}} \\
        & \textbf{en-de} & \textbf{en-es} & \textbf{ja-zh} & \textbf{Avg(ESA)} & \textbf{en-de} & \textbf{en-es} & \textbf{ja-zh} & \textbf{Avg(ESA)} \\
        \midrule
        DA only & 50.41 & 68.51 & 54.48 & 54.72 & 85.62 & \textbf{82.13} & 92.62 & 87.65 \\
        MQM only & 54.71 & 68.80 & 56.36 & 54.20 & 85.55 & 78.56 & 89.94 & 86.45 \\
        DA + MQM & 54.71 & 68.92 & 56.16 & 54.91 & 85.21 & 78.89 & 90.67 & \textbf{87.80} \\
        \midrule
        DA\textrightarrow MQM & 55.40 & 69.11 & 57.90 & 55.00 & 85.91 & 78.12 & \textbf{93.64} & 85.74 \\
        DA\textrightarrow (DA + MQM) & \textbf{55.66} & \textbf{69.25} & \textbf{58.24} & \textbf{55.14} & \textbf{86.60} & 77.92 & 92.88 & 87.08 \\
        \bottomrule
    \end{tabular}
    \vspace{-0.05in}
    \caption{Meta-evaluation scores of reference-based models (non-hybrid) on the WMT24 validation set, using a variety of one- and two-stage fine-tuning protocols, with the \textrightarrow\ symbol indicating two stages. ``DA + MQM'' denotes the combination of DA and MQM scores, with a score type indication provided in the input. The correlation scores are shown for all 3 MQM language pairs individually, and for the 9 ESA language pairs averaged.}
    \label{tab:metricx_results_training_protocols}
\end{table*}

We start by examining the effects of mixing DA and MQM training data, together with using the score type indicators in the input.
As a reminder, raw DA scores are used to train the model to predict ESA scores, since they both use the same scale and follow similar distributions.
The first 3 rows of Table~\ref{tab:metricx_results_training_protocols} compare the combined fine-tuning with fine-tuning on DA data only and MQM data only. We can see that the ``DA + MQM'' model (row~3) performs on par with the ``MQM only'' model (row~2) on the MQM validation sets and on par with the ``DA only'' model (row~1) on the ESA validations sets.
This demonstrates that the model can effectively learn to predict both types of scores without sacrificing performance in either of them.

Our next observation is that simple two-stage fine-tuning (DA first then MQM) also achieves this goal, except for the system-level performance on ESA, which is around 2 points behind both the ``DA + MQM'' and the ``DA only'' model (compare row~4 against rows~3 and~1).
Finally, we show that by combining two-stage fine-tuning and DA/MQM data mixing (row~5), we significantly boost the system-level performance on the ESA sets, while maintaining or further improving the performance on the MQM sets, as well as maintaining the segment-level performance.

\subsubsection{MetricX-25 Submissions}

\begin{table*}
    \small
    \centering
    \begin{tabular}{>{\centering\arraybackslash} m{0.18\linewidth}  >{\centering\arraybackslash} m{0.07\linewidth} >{\centering\arraybackslash} m{0.07\linewidth} >{\centering\arraybackslash} m{0.07\linewidth} >{\centering\arraybackslash} m{0.08\linewidth} >{\centering\arraybackslash} m{0.07\linewidth} >{\centering\arraybackslash} m{0.07\linewidth} >{\centering\arraybackslash} m{0.07\linewidth} >{\centering\arraybackslash} m{0.08\linewidth}}
        \toprule
        \multirow{2}{*}{\textbf{\makecell{MetricX variant}}} & \multicolumn{4}{c}{\textbf{Segment-level pairwise accuracy}} & \multicolumn{4}{c}{\textbf{System-level soft pairwise accuracy}} \\
        & \textbf{en-de} & \textbf{en-es} & \textbf{ja-zh} & \textbf{Avg(ESA)} & \textbf{en-de} & \textbf{en-es} & \textbf{ja-zh} & \textbf{Avg(ESA)} \\
        \midrule
        24-Hybrid-QE & 52.60 & 68.50 & 53.00 & -- & \textbf{87.80} & 78.90 & 87.50 & -- \\
        24-Hybrid & 53.20 & 68.50 & 53.90 & -- & 87.40 & \textbf{79.90} & 89.70 & -- \\
        \midrule
        25-QE\textsuperscript{\textdagger} & 54.97 & \textbf{69.42} & 57.21 & 54.34 & 85.45 & 78.29 & 91.34 & 84.91 \\
        25-Ref\textsuperscript{\textdagger} & \textbf{55.66} & 69.25 & \textbf{58.24} & \textbf{55.14} & 86.60 & 77.92 & \textbf{92.88} & 87.08 \\
        25-Hybrid-QE & 54.83 & 69.31 & 56.66 & 54.11 & 85.42 & 77.74 & 91.59 & 86.32 \\
        25-Hybrid* & 55.45 & 69.14 & 57.72 & 54.87 & 85.82 & 77.00 & 92.00 & \textbf{87.61} \\
        \bottomrule
    \end{tabular}
    \vspace{-0.05in}
    \caption{Performance of our MetricX-25 submissions compared to the MetricX-24 baseline on WMT24. ``Hybrid`` and ``Hybrid-QE'' rows correspond to the same model, only evaluated with and without references, respectively. The correlation scores are shown for all 3 MQM language pairs individually, and for the 9 ESA language pairs averaged. *Primary submission. \textsuperscript{\textdagger}Secondary submissions.}
    \label{tab:metricx_results_final}
    \vspace{-0.1in}
\end{table*}

Table~\ref{tab:metricx_results_final} summarizes our MetricX-25 submissions to the quality score prediction task and compares them against MetricX-24 -- one of the three top-performing systems in last year's shared task -- as a baseline.
The submissions only differ in the combination of examples they were trained on (with or without source/reference) and thus their expected input: the primary submission is a hybrid system, whereas the two secondary submissions are a purely \gls*{qe} and a purely reference-based system, respectively.

As the table shows, all the MetricX-25 submissions (rows 3--6) significantly outperform the MetricX-24 baseline (rows 1--2) at the segment level, but the results are mixed at the system level.
The ja-zh language pair exhibits by far the largest improvement, suggesting that Gemma 3 has a stronger understanding of Japanese and/or Chinese than mT5.
Our expectation is that this is true for many more languages, making MetricX-25 a more robust automatic evaluation metric than its mT5-based predecessors.

We chose the hybrid model as our primary submission, despite being slightly outperformed by the reference-based variant (compare rows 6 and 4), because of its input flexibility.
Since the official test set consists of language pairs both with and without references available, the reference-based model would likely perform poorly on the latter.
Moreover, the majority of the challenge sets also do not provide references, so MetricX-25-Ref would not be able to participate in them.

\section{Error Span Detection: \textit{GemSpanEval}}
\label{sec:gemspaneval}

In this section, we describe our submission to subtask~2: Error Span Detection.
We denote our system \textit{GemSpanEval}, as it is a span-level prediction model based on Gemma~3.

\subsection{Overview}

Last year's shared task on error span detection~\cite{zerva-etal-2024-findings}, despite low participation, showed that span-level error prediction remains a challenging task.
Specifically, \citet{shan-etal-2024-hw} found that generative methods based on LLMs, such as GPT-4o-mini~\cite{hurst2024gpt} or Tower-Instruct-7B~\cite{tower_llm_2024}, still lag behind encoder-only models like \textsc{CometKiwi}~\cite{rei-etal-2022-cometkiwi}. 
Our shared task submission aims to explore how far we can push generative models at the error span detection task when using a recent strong multilingual open-weights model, in our case Gemma~3 27B, fine-tuned for the error span detection task.

\subsection{Error Span Detection}

We adapt the AutoMQM setup~\cite{fernandes-etal-2023-devil} by predicting MQM error spans in JSON format similar to \citet{finkelstein2024jack}.
LLMs tend to be good at producing valid JSON output, making parsing the structured output straightforward.
Each error contains the text span of the machine translation or the source segment (for omissions) along with a severity and a category.
Note that for the error span detection task the category and source errors, such as omissions, are not used.

While models are able to identify and extract spans, we also need to find the spans in the original text for this task.
A simple way to do so is to perform a string search.
This is successful for most error spans, as they are unique substrings of the source or the machine translation.
However, there is also a considerable portion of error spans that are not unique, often related to punctuation or short frequent words.
For example, in the WMT24 en-de MQM data, 21\% of error spans are not unique.
Note though, that the problem is less pronounced for the shared task evaluation, as it is based on the character F1 score, to which short spans contribute less than long spans.

\begin{figure}[ht]
\centering
\begin{lstlisting}
English source:
```I have not made use of the timer, preferring to turn them on and off myself. I can see this feature as useful in an office setting with houseplants or if on vacation```
German machine translation:
```Ich benutze den T|\textcolor{red}{im}|er nicht, sondern schalte |\textbf{ihn}| lieber selbst ein und aus. Ich sehe diese Funktion als |\underline{nützlich }||\underline{\textbf{im}}||\underline{ Büro}| |\textbf{mit}| Z|\textcolor{red}{im}|merpflanzen oder |\textcolor{red}{im}| Urlaub.```
\end{lstlisting}
\textbf{Response}:
\begin{lstlisting}
[
  {"span": "|\textbf{im}|", "span_with_context": "|\underline{nützlich im Büro}|", ...},
  {"span": "|\textbf{ihn}|", ...},
  {"span": "|\textbf{mit}|", ...}
]
\end{lstlisting}
\vspace{-0.1in}
\caption{Example translation with non-unique error spans, where span context text is included.}
\label{fig:json_prompt_response}
\end{figure}

To be able to uniquely identify short spans, we modify the model response to include additional context for any span that is not unique.
We expand the context to the previous and next word by looking for the previous and next space character.
For Chinese and Japanese, we extend the context by one character at a time.
The context is extended until we find a unique substring.
See Figure~\ref{fig:json_prompt_response} for an example of translation error spans with context in JSON format, and Figure~\ref{fig:json_prompt_response_full} in the Appendix for the full prompt and output.
The rest of the prompt and format follows~\citet{finkelstein2024jack}.
We do not use ICL examples, as we train a dedicated model.
All data is presented twice: once with the reference translation and once without.
This allows us to evaluate the model as a \gls*{qe} and as a reference-based model both.

\subsection{Experimental Setup}
For development, we train on \gls*{mqm} data from WMT20--23 and evaluate on WMT24 MQM data.
We evaluate using character-level F1 score, which was used as the shared task metric in previous years~\cite{zerva-etal-2024-findings} and also the current year.
The metric takes into account error severities and gives partial credit for predicting an error span with the wrong error severity.
For the submission system, we include the en-de and ja-zh WMT24 MQM data in training as well, holding out en-es for evaluation.
We include the latest data in order to fine-tune the model on longer segments too.
Before WMT24, only WMT23 en-de provided paragraph-level data, while all other data is at the sentence level.
Additionally, we hope to increase the coverage of translation errors of modern LLM-based translation models.

We fine-tune a 27B Gemma 3 model using Kauldron SFT tooling.\footnote{\url{https://kauldron.readthedocs.io/en/latest/}}
We use the Adafactor~\cite{shazeer2018adafactor} optimizer with a learning rate of 0.0001 and a batch size of 64, running for 20K steps, which covers a little under 2 epochs of the training data.
As baselines we report \textsc{xComet}-XXL~\cite{guerreiro-etal-2024-xcomet} scores.
\textsc{xComet} is a strong encoder-only model that was trained to rate segments and also label translation error token spans.
Additionally, we evaluate Gemma~3 27B without fine-tuning, prompting it to produce JSON error spans, as shown in Figure~\ref{fig:json_prompt_response_full}.
Note that this baseline does not include the span context for short spans.

\subsection{Results and Submission Details}

The experimental results for the span-level error prediction task are shown in Table~\ref{tab:span_level_f1_scores}.
GemSpanEval-QE~v1 denotes an initial training run for just 10K steps and without span context for short spans.
GemSpanEval-QE and GemSpanEval are the \gls*{qe} and reference-based evaluations, respectively, of a model trained without WMT24 data.
The final row adds WMT24 en-de and ja-zh data.
Therefore, for the last row, we should only take the results as an indication of how much of WMT24 has been memorized, not of how good the model is, except for en-es.
The final shared task submission is based on the model that was trained \emph{with} WMT24 data.
The primary submission uses the reference while the secondary submission is reference-free.
Both submissions use the same model and just differ in whether references are shown at inference time.
Note that, when preparing the submission data, we ran into model repetition problems that resulted in invalid JSON output.
For these cases we fell back to the original model, GemSpanEval-QE~v1.
For the shared task submission we find 22\% of errors spans to not be unique, while this corresponds to only 5\% of the characters of error spans, as non-unique spans tend to be short.
Consequently, for WMT24 en-de this leads to a marginal F1 improvement of 0.08.

\begin{table}[ht]
    \centering
    \small
    \setlength{\tabcolsep}{3pt}
    \begin{tabular}{lcccc}
        \toprule
        \textbf{System} & \textbf{en-de} & \textbf{en-es} & \textbf{ja-zh} & \textbf{Avg.} \\
        \midrule
        \textsc{xComet}-XXL-QE & 24.28 & 10.11 & 14.30 & 16.23 \\
        \textsc{xComet}-XXL & \textbf{25.43} & 11.02 & 24.94 & \textbf{20.46} \\
        Gemma 3 27B & 17.94 & 8.19 & \textbf{28.42} & 18.18 \\
        \midrule
        GemSpanEval-QE v1 & 17.51 & 14.43 & 22.75 & 18.23 \\
        GemSpanEval-QE & 20.85 & 13.06 & 24.72 & 19.54 \\
        GemSpanEval & 21.79 & 13.73 & 25.28 & 20.27 \\
        + WMT24 train &  \textcolor{gray}{27.26} & \textbf{14.37} & \textcolor{gray}{37.09} & \textcolor{gray}{26.24} \\
        \bottomrule
    \end{tabular}
    \caption{WMT24 character level F1 scores for the error span prediction task. Numbers where we train on the development set are grayed out but kept for reference.}
    \label{tab:span_level_f1_scores}
    \vspace{-0.1in}
\end{table}

From the experimental results during development in Table~\ref{tab:span_level_f1_scores}, we see that the encoder-only \textsc{xComet}-XXL model is a strong baseline showing the best result for en-de. 
The Gemma~3 baseline that was not fine-tuned also shows generally good performance, even achieving the best result of all evaluated systems for ja-zh, better than the fine-tuned model.
This is likely due to the task being difficult and highly dependent on rater behavior, which varies significantly across years and languages.
Using references consistently achieves higher character F1 across all three language pairs.
Surprisingly, the model trained with WMT24, while showing the best results in terms of character F1, still shows a significant gap, despite seeing the test data during training for en-de and ja-zh.
This shows that $\sim$2 epochs in the current training setup was not enough to completely memorize the training data.
For the held-out language, en-es, we see the best score across all settings, leading to the decision to use the model trained with WMT24 as our submission to the shared task.

\todo{maybe report results with and without the unique span matching}

\section{Related Work}
\label{sec:related_work}

For decades, right until the recent advent of LLMs, the most widely adopted automatic evaluation metrics would express the predicted machine translation quality as a scalar score.
This is the case for metrics ranging from simple lexical overlap metrics, such as BLEU~\citep{papineni-etal-2002-bleu} and ChrF~\citep{popovic-2015-chrf}, to learned metrics including \textsc{Bleurt}~\citep{sellam-etal-2020-bleurt,pu-etal-2021-learning}, \textsc{Comet}~\citep{rei-etal-2020-comet,rei-etal-2022-comet} and MetricX~\citep{juraska-etal-2023-metricx,juraska-etal-2024-metricx}.
The feasibility of prompting general-purpose LLMs to score translations has also been studied~\citep{kocmi-federmann-2023-large,leiter-etal-2023-eval4nlp,leiter-eger-2024-prexme} and this approach has been shown to be competitive with fine-tuned dedicated models, especially in system-level performance.
Among recent methods, there has been an increasing proportion of metrics providing structured \citep{perrella-etal-2022-matese,kocmi-federmann-2023-gemba,fernandes-etal-2023-devil,guerreiro-etal-2024-xcomet} or natural language explanations \citep{xu-etal-2023-instructscore} for the predicted scores, most of which are based on LLMs.

Since the era of lexical metrics, which require one or more reference translations to evaluate a machine translation, metrics have evolved to rely increasingly more on the source segment~\citep{rei-etal-2020-comet,rei-etal-2022-comet,juraska-etal-2024-metricx}.
This is enabled by the multilingual pretrained models they are typically built on top, such as XLM-R~\citep{conneau-etal-2020-unsupervised} or mT5~\citep{xue-etal-2021-mt5}.
In fact, \emph{reference-free} (or \emph{quality estimation}; QE) metrics do not lag far behind their \emph{reference-based} counterparts anymore, as evidenced by the most recent WMT Metrics shared tasks~\citep{freitag-etal-2023-results,freitag-etal-2024-llms}.
That being said, high-quality references do provide added value to automatic metrics in most cases, helping them make even more accurate quality predictions.
Therefore, metrics nowadays typically employ a unified (or hybrid) input approach, allowing them to make a reference-based prediction whenever a reference is available, and a QE prediction otherwise~\citep{wan-etal-2022-unite,guerreiro-etal-2024-xcomet,juraska-etal-2024-metricx}.
This is the case with our primary MetricX-25 and GemSpanEval submissions as well.

Current approaches to error span annotation are often based on encoder-only models predicting error severities per token.
One such approach is \textsc{CometKiwi}~\cite{rei-etal-2022-cometkiwi,rei-etal-2023-scaling} or, more recently, \textsc{xComet}~\cite{guerreiro-etal-2024-xcomet}.
In last year's Error Span Detection shared task, \textsc{CometKiwi} was shown to be a competitive baseline, coming in ahead of the (single) submitted system~\cite{zerva-etal-2024-findings}.
\citet{kocmi-federmann-2023-gemba} showed that GPT-4 can simply be prompted to produce MQM error spans, denoting the method GEMBA-MQM.
\citet{fernandes-etal-2023-devil} showed that fine-tuning for the generative error span prediction can improve performance compared to prompting LLMs alone.
Recent translation-oriented LLMs, such as Tower~\cite{tower_llm_2024,rei2025tower+} also include generative error span prediction as part of the supported tasks in the training data.

\section{Conclusion}
\label{sec:conclusion}

We introduced MetricX-25 and GemSpanEval, our submissions to the WMT25 Evaluation Shared Task, both built upon the Gemma~3 foundation model, but in very different ways.
We demonstrated that MetricX-25, adapting Gemma~3 to an encoder-only architecture, can be trained to effectively predict ESA and MQM quality scores and significantly outperforms its predecessor, MetricX-24, in segment-level performance.
For error span detection, our generative model, GemSpanEval, proved to be competitive with a strong sequence-tagging baseline.
Additionally, we showed how error span context can be used to identify unique error spans.
Our work demonstrates that a strong multilingual foundation model, such as Gemma~3, can successfully be used for both regression-based and generative translation evaluation metrics.

\bibliography{anthology.2015,custom}
\bibliographystyle{acl_natbib}

\clearpage

\appendix
\section{MetricX-25 Implementation Details}
\label{app_sec:metricx_implementation_details}

With the input context length of 4096 SPM tokens, each training run requires 64 TPUs.
Using a batch size of 128, we train our models for 10K steps in the first stage, using a learning rate of 0.00005 with a cosine decay following 100 steps of linear warm-up.
We then fine-tune the best checkpoint for another 10K steps in the second stage, lowering the peak learning rate to 0.00001.

\begin{figure*}[ht]
\centering
\textbf{Prompt}
\begin{lstlisting}
You are an annotator for the quality of machine translation. Your task is to identify errors and assess the quality of the translation.
Based on the source segment, human-generated reference translation, and machine translation surrounded with triple backticks, identify error types in the translation and classify them. The categories of errors are: accuracy (addition, mistranslation, omission, untranslated text), fluency (character encoding, grammar, inconsistency, punctuation, register, spelling), style (awkward), terminology (inappropriate for context, inconsistent use), non-translation, other, or no-error.
Each error is classified as one of three severities: critical, major, and minor. Critical errors inhibit comprehension of the text. Major errors disrupt the flow, but what the text is trying to say is still understandable. Minor errors are technically errors, but do not disrupt the flow or hinder comprehension.

Make sure your response is a strict and valid json object that could be parsed with json.loads() in python.

English source:
```The lights are dimmable, but I use the strongest setting only. I have not made use of the timer, preferring to turn them on and off myself. I can see this feature as useful in an office setting with houseplants or if on vacation```
German machine translation:
```Die Lichter sind dimmbar, aber ich benutze nur die stärkste Einstellung. Ich benutze den Timer nicht, sondern schalte |\textbf{ihn}| lieber selbst ein und aus. Ich sehe diese Funktion als |\underline{nützlich }||\underline{\textbf{im}}||\underline{ Büro}| |\textbf{mit}| Zimmerpflanzen oder im Urlaub.```
\end{lstlisting}
\textbf{Response}:
\begin{lstlisting}
[
  {"span": "|\textbf{im}|", "severity": "minor", "category": "accuracy/mistranslation", "span_with_context": "|\underline{nützlich im Büro}|"},
  {"span": "|\textbf{ihn}|", "severity": "minor", "category": "accuracy/mistranslation"},
  {"span": "|\textbf{mit}|", "severity": "minor", "category": "accuracy/mistranslation"}
]
\end{lstlisting}
\caption{Example prompt and response for AutoMQM error span identification. We omit the error span attribute \texttt{is\_source\_error} for brevity. Each span that is not unique receives an additional attribute \texttt{span\_with\_context}.}
\label{fig:json_prompt_response_full}
\end{figure*}

\end{document}